\documentclass[a4paper,11pt,twocolumn,twoside]{article}
\usepackage{sepln}
\usepackage{fullname}
\usepackage[utf8]{inputenc}
\usepackage{booktabs}
\usepackage{graphicx}
\usepackage{amsmath}
\usepackage{amssymb}
\usepackage{xcolor}
\usepackage[english]{babel}
\usepackage{url}

\input epsf
\renewcommand{\textcolor}[2]{#2} 

\title{\textit{You Are What You Prompt:} Prompt Quality, Domain Shift, and Uncertainty in Agrifood Vision-Language Models}

\author {\textbf{Andrea Morales-Garzón$^1$} \textbf{Salvador Lopez-Joya$^1$} \textbf{Miguel López-Pérez$^1$} \\\textbf{Maria J. Martin-Bautista$^1$}\\
$^1$Dept. Computer Science and Artificial Intelligence, University of Granada\\
amoralesg@decsai.ugr.es, slopezjoya@ugr.es, \{mlopez,mbautis\}@decsai.ugr.es\\
}

\seplntranstitle{Calidad de prompts, cambio de dominio e incertidumbre en modelos visión-lenguaje agroalimentarios}

\seplnclave{Clasificación zero-shot, Modelos 
de visión y lenguaje, Ingeniería de prompts, 
PLN específico de dominio, 
Cuantificación de incertidumbre}

\seplnresumen{
Los modelos de visión y lenguaje permiten la clasificación zero-shot mediante 
descripciones en lenguaje natural, pero su rendimiento es sensible a la formulación de los prompts, especialmente en dominios 
especializados. El método Zero-shot Prompt Ensembling (ZPE) aborda esto ponderando los prompts según su señal discriminativa, aunque su comportamiento bajo cambio de dominio permanece inexplorado. Evaluamos ZPE en el dominio agroalimentario usando CLIP y SigLIP sobre cuatro conjuntos de datos y cuatro colecciones de prompts, cubriendo benchmarks de comida (in-distribution, ID) y agricultura (out-of-distribution). ZPE aporta beneficios limitados en condiciones ID, pero mejora sustancialmente el rendimiento y la calibración bajo cambio de dominio, donde colecciones de prompts específicas de dominio (51--52 prompts) superan consistentemente a las genéricas (247--426). El análisis léxico muestra que ZPE actúa como un detector no supervisado de alineación de dominio sin acceso a etiquetas. Además, introducimos PID (Prompt-based Inconsistency Detection), que reutiliza el desacuerdo entre prompts como señal de incertidumbre epistémica, mejorando la detección de fallos bajo cambio de dominio severo donde las medidas de confianza estándar colapsan.
}

\seplnkey{Zero-shot Classification, Visual Language models, Prompt Engineering, Domain Specific NLP, Uncertainty Quantification}

\seplnabstract{Vision--language models enable zero-shot classification through natural language 
prompts, but performance is sensitive to prompt formulation, especially in specialized domains. Zero-shot Prompt Ensembling (ZPE) addresses this by weighting prompts by discriminative signal, yet its behavior under domain shift remains unexplored. We evaluate ZPE in the agrifood domain using CLIP and SigLIP across four datasets and four prompt pools, spanning in-distribution (ID) food and out-of-distribution agricultural benchmarks. ZPE provides limited benefit under ID conditions but substantially improves performance and calibration under domain shift, where domain-specific pools of 51--52 prompts consistently outperform generic pools of 247--426. Lexical analysis shows that ZPE acts as an unsupervised domain-alignment detector without label access. We further introduce PID (Prompt-based Inconsistency Detection), which repurposes prompt disagreement as epistemic uncertainty, improving failure detection under severe domain shift where standard confidence measures collapse.}

\firstpageno{1}

\begin{document}


\setlength\titlebox{22cm} 

\label{firstpage} \maketitle

%

\section{Introduction}
\label{sec:intro}

Vision–language models (VLMs) such as CLIP~\cite{radford2021learning} and SigLIP \cite{zhai2023sigmoid} enable zero-shot classification by aligning images with natural language descriptions of classes. In this setting, the formulation of text prompts plays a critical role in model performance. A widely adopted strategy to improve robustness is \emph{prompt ensembling}~\cite{liu2023pre}, where multiple prompt templates are used to predict each class and their outputs are aggregated. The underlying assumption is that a diverse prompt pool can mitigate variability introduced by weaker individual prompts~\cite{pitis2023boosted}.

In practice, prompt ensembling is not always beneficial. When VLMs trained on general-purpose data are applied to specialized domains, prompt engineering becomes substantially more challenging. In preliminary experiments, it has been observed that adding more prompts can degrade performance, particularly when the pool contains poorly aligned or noisy templates~\cite{morales2026zero}. This raises a fundamental question: \emph{how can we make prompt ensembling robust to prompt quality, and which prompts deserve to be trusted?}

Zero-shot Prompt Ensembling (ZPE) addresses this by assigning higher weights to prompts with stronger discriminative signal, estimated without any external validation set \cite{allingham2023simple}. While ZPE has been shown to improve accuracy over uniform ensembling in general settings, its behavior in domain-shifted scenarios remains underexplored. In particular, it is unclear whether the disagreement signal that drives ZPE carries meaningful information when the model operates far from its training distribution.

In this study, we investigate these questions across specific agrifood domains, including food classification and agriculture-related tasks such as plant disease recognition. 
\textcolor{blue}{We treat food-related datasets as \emph{in-distribution} (ID) proxies based on two observations: (i) CLIP's zero-shot accuracy on food datasets such as Food-101 approaches fully supervised performance (Radford et al., 2021), indicating strong coverage of food imagery in web-crawled pretraining corpora; and (ii) platforms heavily represented in these corpora (e.g., Pinterest, Shopify) are dominated by food and consumer content, whereas agricultural pathology imagery remains confined to specialist repositories, making it naturally \emph{out-of-distribution} (OOD) for these models. Our own results corroborate this, as we
observe robust performance in food-domain benchmarks but notable disparities in OOD agricultural and bean disease datasets, highlighting the sensitivity of VLMs to domain-specific vocabulary and fine-grained visual variation.
Our empirical study reveals a nuanced picture: ZPE provides limited benefit under ID conditions, where uniform ensembling already performs well, but becomes substantially more useful under OOD conditions, where domain shift is severe.} 

Building on this finding, we introduce \textbf{PID} (Prompt-based Inconsistency Detection), a method that repurposes prompt disagreement within the ZPE ensemble as a proxy for epistemic uncertainty. PID serves as a failure detector that improves model reliability under domain shift without modifying the prediction pipeline or incurring additional computational cost. By gating predictions through disagreement-based uncertainty estimates, PID offers a principled way to flag unreliable outputs precisely in the OOD regime where standard confidence measures tend to break down.

The contributions of this work are:
\begin{itemize}
    \item We evaluate ZPE across multiple agrifood domains under both ID and OOD conditions, showing that its benefit is regime-dependent: marginal under ID but substantial under OOD data.
    \item We identify consistent performance disparities in OOD settings, particularly in agriculture and bean disease benchmarks, and contrast these with robust ID results in food classification.
    \item We introduce PID, a prompt disagreement measure that acts as a quality-gated epistemic uncertainty signal within the ZPE framework, and demonstrate its effectiveness as a reliability proxy for OOD selective prediction.
    \item We analyze from a lexical perspective the prompt quality, showing that terms referring to concrete visual content are associated with higher ZPE scores, while more abstract or weakly visual words tend to correspond to lower-quality prompts.
\end{itemize}

\section{Related Work}
\label{sec:related}

Recent progress in VLMs, particularly since the introduction of CLIP \cite{radford2021learning}, has shown that strong zero-shot performance depends not only on the pretrained model, but also on how we interact with it through language. This has motivated a growing line of work exploring how to design, select, and aggregate prompts, as well as how to assess the reliability of their predictions. In this section, we review prior research on prompt engineering for VLMs, calibration in zero-shot VLMs, agrifood image classification, and uncertainty estimation in ensembles.

\paragraph{Prompt engineering for VLMs.}

Prompt design plays a critical role in zero-shot performance, as different
formulations of the same class can lead to different predictions.
Menon \& Vondrick proposed classification by description, replacing class-name
prompts with a set of descriptors automatically generated with an LLM. Instead
of querying ``a photo of a tiger'', the model checks for discriminative features
(stripes, claws, size). This produces better accuracy and interpretability,
since each classification decision is grounded in verifiable visual
attributes~\cite{menon2022visual}. Learned prompt tuning methods such as CoOp
and CoCoOp~\cite{zhou2022coop,zhou2022conditional} replace fixed templates with
learnable context vectors, improving accuracy on seen classes at the cost of
requiring labeled data and reduced zero-shot generalizability --- limitations
that ZPE avoids by operating entirely without supervision. ZPE assigns weights to prompts in a large pool
based on their discriminative signal and produces a weighted average ensemble
prediction~\cite{allingham2023simple}. In our work, we build on ZPE and extend
it to the agrifood domain, analyzing prompt quality from a lexical perspective
and using prompt disagreement as an uncertainty signal for failure detection.

\paragraph{Calibration in zero-shot models.}

Calibration has been widely studied in modern neural networks, where it is
often observed that high accuracy does not necessarily imply well-calibrated
predictions. Deep neural networks are shown to be systematically overconfident,
a phenomenon that becomes more pronounced with increased model depth, width,
batch normalisation, and weight decay~\cite{guo2017calibration}. Temperature
scaling is introduced as a simple yet highly effective post-hoc method that
rescales logits with a single parameter, improving calibration without affecting
classification accuracy. More recent work has revisited these findings in the
context of large-scale vision models. Minderer et al.\ evaluated several
architectures, including Vision Transformers, MLP-Mixers, and large pretrained
models such as CLIP~\cite{minderer2021revisiting}, showing that modern
architectures are often significantly better calibrated and that calibration can
even improve with model scale under distribution shift. In our work, we do not
perform explicit calibration, but we show that prompt disagreement within ZPE
correlates with unreliable predictions under domain shift, offering a
complementary and label-free reliability signal.

\paragraph{Agrifood image classification.}
Recent work has begun to explore the application of vision-language models in agricultural and food-related domains, where fine-grained recognition, domain shift, and limited supervision pose significant challenges. Studies such as AgriCLIP \cite{nawaz2025agriclip} and \cite{kaur2025visual} show that although CLIP and BLIP-2 \cite{li2023blip} provide strong baselines, their performance degrades significantly in tasks requiring fine-grained agricultural expertise, environmental robustness, or reasoning over contextual information. Similarly, E-CLIP \cite{agriculture15111173} and \cite{imeraj2025clip} propose hybrid architectures that combine CLIP embeddings with detection heads or structured botanical features to improve robustness in crop and plant disease recognition.

To address the lack of domain-specific data, several multimodal datasets have been proposed, including VL-PAW \cite{yu2025vl}, AgriCoT \cite{wen2025agricot}, WisWheat \cite{yuan2025wiswheat}, and AgMMU \cite{gauba2025agmmu}, which introduce image--text pairs, reasoning tasks, and expert-level annotations for agricultural decision-making. 

For evaluation, Food-101 \cite{bossard2014food} remains the standard benchmark for fine-grained food recognition, comprising 101 categories with substantial intra-class variability due to diverse real-world imaging conditions. Despite its relative simplicity, it serves as a key reference for vision-language models. Beans, for instance, is a three-class plant disease dataset (angular leaf spot, bean rust, and healthy) with a pronounced domain gap, as its labels require specialized botanical knowledge largely absent from general-purpose pretraining. As a result, zero-shot models often perform close to chance, making it a useful stress test for robustness under distribution shift. 

\paragraph{Epistemic uncertainty in ensembles.}
Ensemble disagreement has been established as a practical proxy for epistemic uncertainty, capturing the model's lack of confidence due to 
insufficient or mismatched training signal~\cite{lakshminarayanan2017simple}. The diversity among ensemble members is key to this property: more diverse predictions correlate with higher uncertainty, particularly under distribution shift~\cite{fort2019deep}. Critically, ensemble-based uncertainty estimates are known to degrade gracefully but detectably under dataset shift, making disagreement a reliable signal precisely in the OOD regime~\cite{ovadia2019can}. In the context of prompt ensembling, we interpret disagreement across prompt predictions as an indicator of epistemic uncertainty under domain shift: when a model operates far from its training distribution, prompts probing the same class from different linguistic angles are less likely to agree, producing a detectable inconsistency signal. PID operationalizes this intuition directly, without requiring additional training or distributional assumptions.

\section{Methodology}
\label{sec:method}

\subsection{Zero-Shot classification}

Given an image $x$ and a set of class names $\{c_1,\ldots,c_C\}$, a
VLM encodes the image into $\mathbf{v}\!=\!f_\theta(x)\!\in\!\mathbb{R}^D$
and each text prompt into
$\mathbf{t}_{p,c}\!=\!g_\phi(\mathrm{tmpl}_p(c))\!\in\!\mathbb{R}^D$.
The logit of class $c$ under prompt $p$ is:
\begin{equation}
  L^{(p)}_c(x) = \alpha \cdot
  \langle \mathbf{v}(x),\, \mathbf{t}_{p,c} \rangle,
  \label{eq:logit}
\end{equation}
where $\alpha$ is a learned logit scale, and the single-prompt prediction
is $\hat{c}^{(p)}\!=\!\arg\max_c L^{(p)}_c(x)$.
We evaluate two VLM families:
CLIP~\cite{radford2021learning} uses a contrastive loss ($\alpha\!\approx\!100$);
SigLIP~\cite{zhai2023sigmoid} uses a sigmoid loss ($\alpha\!\approx\!117$),
removing the dependence on batch normalisation.

\subsection{Zero-Shot Prompt Ensembling (ZPE)}
\label{sec:zpe}

Prompt choice significantly impacts zero-shot accuracy, yet practitioners
often select templates arbitrarily.
In \cite{allingham2023simple}, the authors proposed ZPE, a self-supervised weighting
scheme that assigns higher importance to prompts with stronger
discriminative signal, without requiring any labels.

Given a pool of $P$ prompts evaluated over $N$ test images, ZPE computes a scalar quality score per prompt over the full test set, then uses those scores to weight the ensemble.
This is a \textbf{transductive} procedure: all $N$ images must be seen before any individual prediction is made.

\paragraph{Prompt scoring.}
Two scoring variants are proposed~\cite{allingham2023simple}:

\smallskip
\noindent\textbf{(1) Raw} (Algorithm~1). The score is the mean maximum logit
across the test set:
\begin{equation}
  s^{(p)}_\mathrm{raw} =
  \frac{1}{N}\sum_{i=1}^{N} \max_c\; L^{(p)}_c(x_i).
  \label{eq:score_raw}
\end{equation}

\noindent\textbf{(2) Normalized} (Algorithm~2). Logits are first centred by
subtracting the expected logit under the data distribution,
approximated by the test-set mean:
\begin{equation}
  s^{(p)}_\mathrm{norm} =
  \frac{1}{N}\!\sum_{i=1}^{N}
  \max_c \bigl(L^{(p)}_c(x_i) - \hat{\mu}^{(p)}_c\bigr),
  \label{eq:score_norm}
\end{equation}
where $\hat{\mu}^{(p)}_c\!=\!\frac{1}{N}\sum_i L^{(p)}_c(x_i)$ estimates
$\mathbb{E}[L^{(p)}_c]$.
This centring removes the unconditional logit bias of each prompt,
making scores comparable across templates.

\paragraph{Weighted prediction.}
Scores are converted into prompt weights via a softmax with
temperature $\tau\!>\!0$, and the ensemble prediction is:
\begin{equation}
  \hat{c}^{\mathrm{ZPE}} =
  \arg\max_c \sum_{p=1}^{P} w_p \cdot L^{(p)}_c(x),
  \label{eq:ensemble}
\end{equation}
\begin{equation}
  w_p = \frac{e^{s^{(p)}/\tau}}{\sum_{p'} e^{s^{(p')}/\tau}}.
  \label{eq:weights}
\end{equation}
As $\tau\!\to\!\infty$, Eq.~\ref{eq:ensemble} recovers uniform
ensembling; small $\tau$ concentrates weight on the best-scoring prompt.
We select $\tau^*$ by sweeping a grid between 0.1 and 10.0 and choosing the value that maximizes classification accuracy on the held-out validation split. 

\subsection{Prompt-based Inconsistency Detection (PID)}
\label{sec:pid}

ZPE collapses the ensemble into a single point prediction, discarding
the information encoded in prompt \emph{disagreement}.
We propose PID, which repurposes this disagreement as a per-sample
failure detector: given image $x$, how likely is the model to be wrong?

PID adds a confidence score $\kappa(x)$ that enables
\textbf{selective prediction}: predict on high-$\kappa$ samples,
abstain on low-$\kappa$ ones.
It does \emph{not} change the predicted class or its probabilities.

\paragraph{Quality-gated disagreement.}
Let $q_p(x)\!=\!\mathrm{softmax}(L^{(p)}(x))$ be the class distribution
of prompt $p$, and $\mu(x)\!=\!\sum_p w_p\,q_p(x)$ the ZPE-norm
BMA mean.
The quality-gated disagreement is:
\begin{equation}
  D(x) = \sum_{p=1}^{P} w_p^{\mathrm{agr}} \cdot
  \mathrm{KL}\!\bigl(q_p(x) \;\|\; \mu(x)\bigr),
  \label{eq:disagreement}
\end{equation}
where $w_p^{\mathrm{agr}}\!=\!\mathrm{softmax}(s^{(p)}_\mathrm{norm}/\tau_a)$
uses a separate temperature $\tau_a$.
When $\tau_a\!<\!\tau$, only high-scoring prompts contribute to the
disagreement measurement (\emph{quality-gating}).
When $\tau_a\!\to\!\infty$, all prompts contribute equally, recovering
plain mutual information.

\paragraph{Dirichlet concentration.}
The per-sample confidence is modeled as:
$\kappa(x) = \frac{c}{D(x)},$
where $c\!>\!0$ is a scalar.
High $\kappa(x)$: prompts agree $\to$ prediction likely correct.
Low $\kappa(x)$: prompts disagree $\to$ abstain or escalate.

\paragraph{Fitting.}
Both $\tau_a$ and $c$ are fitted on a held-out validation split (20\%
of the evaluation set, sampled before scoring).
$\tau_a^*$ is chosen to maximize AUROC of $\kappa(x)$ as a binary
correctness predictor; $c^*$ minimizes Dirichlet-Categorical NLL.

\paragraph{Role of $\tau_a$.}
The shape of the $\tau_a$ ablation curve is itself a domain diagnostic.
A \emph{peaked} curve (optimal at low $\tau_a$) indicates that
quality-gating is useful: some prompts are reliably better than others.
A \emph{monotonically increasing} curve indicates full domain shift:
no prompt is trustworthy, so collective disagreement across all prompts
is the most informative signal.

\subsection{Calibration Metrics}
We also measure the standard calibration quality of the ensemble:
\begin{itemize}
  \item \textbf{ECE}: average gap between confidence and accuracy in 15
        equal-width bins. It measures how well calibrated the model is.
  \item \textbf{NLL}: computes the negative log-loss. It retains information about how good the predicted probabilities are, $-\frac{1}{N}\sum_n \log p_{\hat{c}}(x_n)$.
  \item \textbf{Brier Score}: it is the mean-squared error between probabilities and real labels. It does not penalize wrong predictions as much as NLL, $\frac{1}{N}\sum_n
        \|\mathbf{p}(x_n) - \mathbf{y}_n\|^2$.
\end{itemize}

\section{Experimental setup}

\subsection{Agrifood datasets}
We evaluate on four benchmarks spanning food and agricultural image classification. All class names are preprocessed to replace underscores with spaces.
This step is critical for beans: naive underscore names degrade accuracy by $\sim$30 percentage points due to tokenization mismatch in the text encoder.

\begin{itemize}
    \item \textbf{Food-101} 101 food categories, 1{,}000 test images per class.
    \item \textbf{Food-11} 11 broad food groups, 1{,}000 samples per class.
    \item \textbf{Agricultural crops} 30-category crop image dataset (829 images), with classes such as rice, maize, or banana, designed for coarse-grained crop classification.
    \item \textbf{Beans} Bean leaf disease classification (angular leaf spot, bean rust, healthy).
\end{itemize}

\subsection{Models}

We evaluate six models spanning two architectures and three scales: SigLIP~\cite{zhai2023sigmoid} (base, SO/400M, large) and CLIP~\cite{radford2021learning} (B/32, B/16, L). 
This selection enables systematic comparison across both architectural paradigms and model capacity, allowing us to assess how design choices in vision-language modeling affect performance on our tasks. 
SigLIP, which introduces sigmoid-based loss as an alternative to softmax-based contrastive learning, represents recent advances in efficient vision-language alignment, while CLIP remains the de facto standard in the field. 
By evaluating models at base, intermediate, and large scales within each architecture, we characterize the scaling behavior and trade-offs between model size and performance across different representation learning frameworks.

\subsection{Prompt pools}

We used four prompt pools: two general ones from the study detailed in~\cite{allingham2023simple}, and two domain-specific pools that we designed to provide visual angles not covered by generic templates. 
Lexical diversity and domain-specific vocabulary varies across pools, motivating the analysis in Section~\ref{sec:lexical}. They are available in the code repository\footnote{Code and supplementary materials are available at:
\url{https://github.com/ugritai/pid_agrifood}.}.
\begin{itemize}
    \item \textbf{Pool-247}: 247 templates with standard photographic and descriptive templates.
    \item \textbf{Pool-426}: it extends Pool-247 with 179 ChatGPT-generated prompts.
    \item \textbf{Pool-Food: (ours, 51 prompts)} templates targeting food appearance, culinary context and texture\textcolor{blue}{, for the Food-101 and Food-11 datasets}. Examples: ``a plate of \{\}.'', ``a close-up of the texture of \{\}.'', ``street food: \{\}.''.
    \item \textbf{Pool-Agri (ours, 52 prompts)}: templates targeting plant parts, disease symptoms, and agronomic context\textcolor{blue}{, designed for the Agriculture and Beans datasets}. Examples: ``a close-up photo of a \{\} leaf.", ``a photo of \{\} showing rust disease.''.
\end{itemize}

\section{Results}

\begin{table}[t]
\centering
\scriptsize
\setlength{\tabcolsep}{3pt}
\caption{Average accuracy across pools using a uniform ensemble.\label{tab:models}}
\resizebox{\columnwidth}{!}{\begin{tabular}{lccccc}
\hline
\textbf{Model} & \textbf{F101} & \textbf{F11} & \textbf{Agri.} & \textbf{Beans} & \textbf{Avg} \\
\hline
CLIP-B/32   & 0.830 & 0.809 & 0.616 & 0.287 & 0.636 \\
CLIP-B/16   & 0.881 & 0.810 & 0.709 & 0.307 & 0.677 \\
CLIP-L/14   & \textbf{0.932} & \textbf{0.848} & \textbf{0.780} & \textbf{0.398} & \textbf{0.740} \\ \hline
SigLIP-B    & 0.912 & \textbf{0.844} & 0.869 & 0.404 & 0.757 \\
SigLIP-L    & 0.943 & 0.845 & 0.910 & \textbf{0.635} & 0.833 \\
SigLIP-400M & \textbf{0.957} & 0.842 & \textbf{0.941} & \textbf{0.635} & \textbf{0.844} \\
\hline
\end{tabular}}
\end{table}

\subsection{Performance and calibration}

\begin{figure*}[t]
    \centering
    \includegraphics[width=1\linewidth]{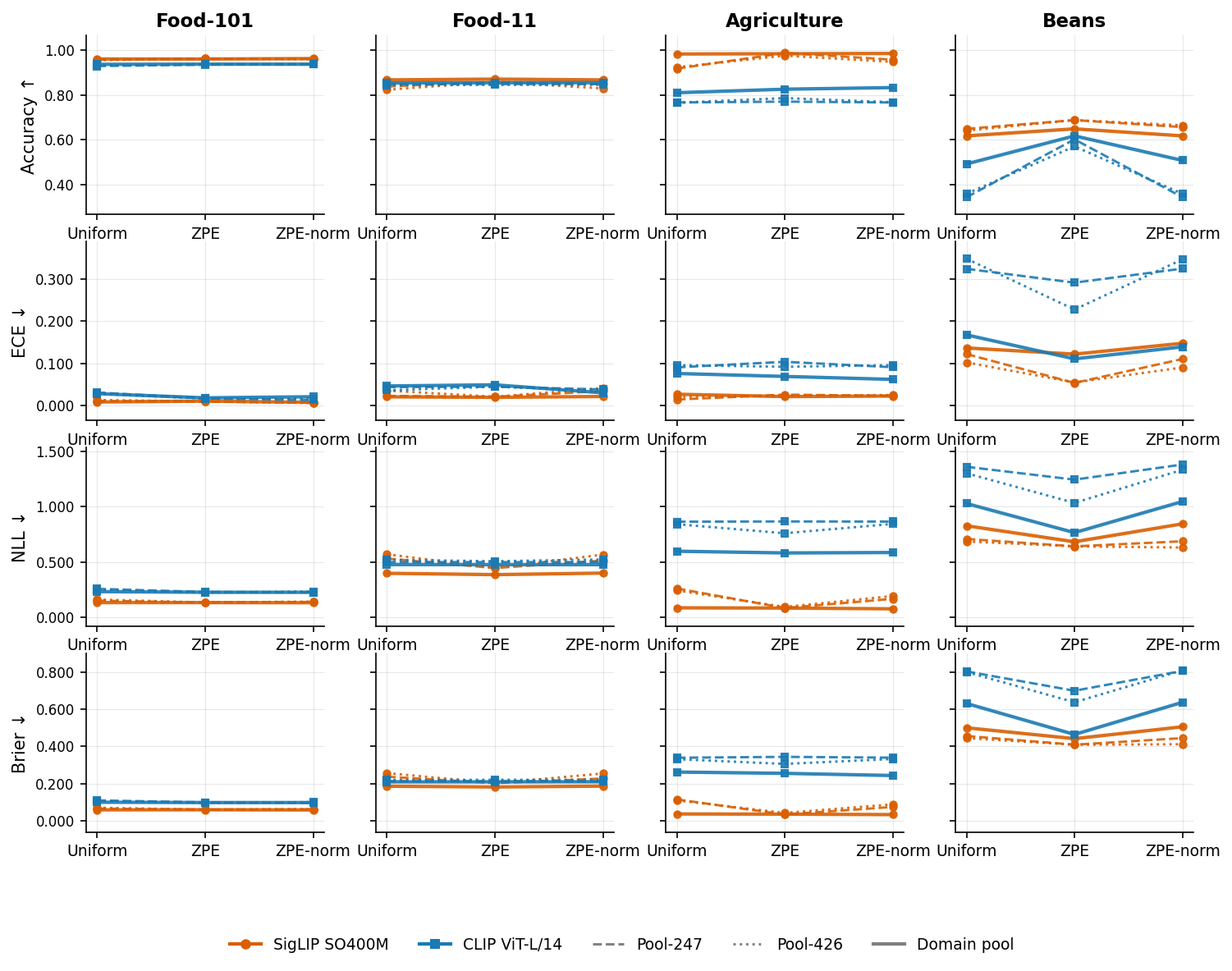}
    \caption{\textcolor{blue}{Zero-shot performance for SigLIP SO400M and CLIP ViT-L/14 (all prompt pools).}}
    \label{fig:calibration_performance}
\end{figure*}

Table~\ref{tab:models} reports the results of the VLM backbones (CLIP and SigLIP) averaging over the different prompt pools in a uniform ensemble. We observe that CLIP-L/14 and SigLIP-400M are consistently the best of their family, respectively. Therefore, to simplify the subsequent performance analysis we only report the results for these  two backbones. 

Fig.~\ref{fig:calibration_performance} shows how performance evolves from uniform prompt ensembling (Uniform) to ZPE-weighted ensembling (ZPE, ZPE-norm) under three prompt pools: Pool-247 (dashed), Pool-426 (dotted), and the domain-specific pool (solid). Rows report accuracy (↑), ECE (↓), NLL (↓), and Brier score (↓); columns correspond to datasets. Overall, SigLIP SO400M (orange) consistently outperforms CLIP ViT-L/14 (blue).

Our evaluation reveals distinct patterns across ID and OOD datasets. On well-aligned ID domains, both architectures achieve high accuracy and perform similarly with minimal calibration error, and their performance remains robust across prompt ensemble strategies. However, on OOD Agriculture domain—proxies for increasing domain shift—both models exhibit higher variation across models and prompt pools, particularly in calibration metrics. In the challenging Beans dataset specifically, we observe dramatic performance degradation and severe miscalibration, with performance becoming highly sensitive to prompt pool and weighting strategy choices. Notably, ZPE-weighted strategies begin to show differentiated behavior in this regime, suggesting that prompt reweighting mechanisms are beneficial when models encounter domain-shifted data. This behavior highlights the critical importance of uncertainty quantification and out-of-distribution detection mechanisms: models maintain well-calibrated confidence on in-distribution data while becoming dangerously overconfident under domain shift. The consistency of these degradation patterns across both SigLIP and CLIP variants—despite SigLIP's superior in-distribution performance—suggests that the bottleneck is fundamentally a domain shift problem rather than an architectural limitation, motivating the need for explicit OOD detection frameworks that can reliably flag when visual inputs deviate significantly from the training distribution.

\subsection{Lexical analysis of prompt quality}\label{sec:lexical}

\paragraph{ZPE score as a signal of prompt quality.}
An important property of ZPE is that prompt scores are computed without using ground-truth labels. Instead, the score reflects how much the ensemble prediction changes when a given prompt is considered with respect to the ensemble mean. To assess whether this signal is meaningful, we compare the ZPE-normalized score of each prompt with its single-prompt accuracy.

An example of this finding is the prompt \textit{``a fundus image with signs of \{\}''}, which ranks among the worst prompts in all four datasets. Its ZPE scores are consistently low (6.4 in Food-101, 2.4 in Food-11, 4.9 in Agriculture, and 0.8 in Beans), and this is matched by low accuracies (0.48, 0.60, 0.60, and 0.37, respectively). This suggests that ZPE is able to identify strongly mismatched prompts without requiring access to labels. This is particularly relevant in agrifood settings, where tasks span heterogeneous visual domains, from plated food images to crop and plant pathology. In this context, prompt pools may contain templates that are semantically plausible but visually mismatched. Here, ZPE is useful to detect prompt quality and discard out-of-scope prompts.

\begin{figure*}[t]
    \centering
    \includegraphics[width=1\linewidth]{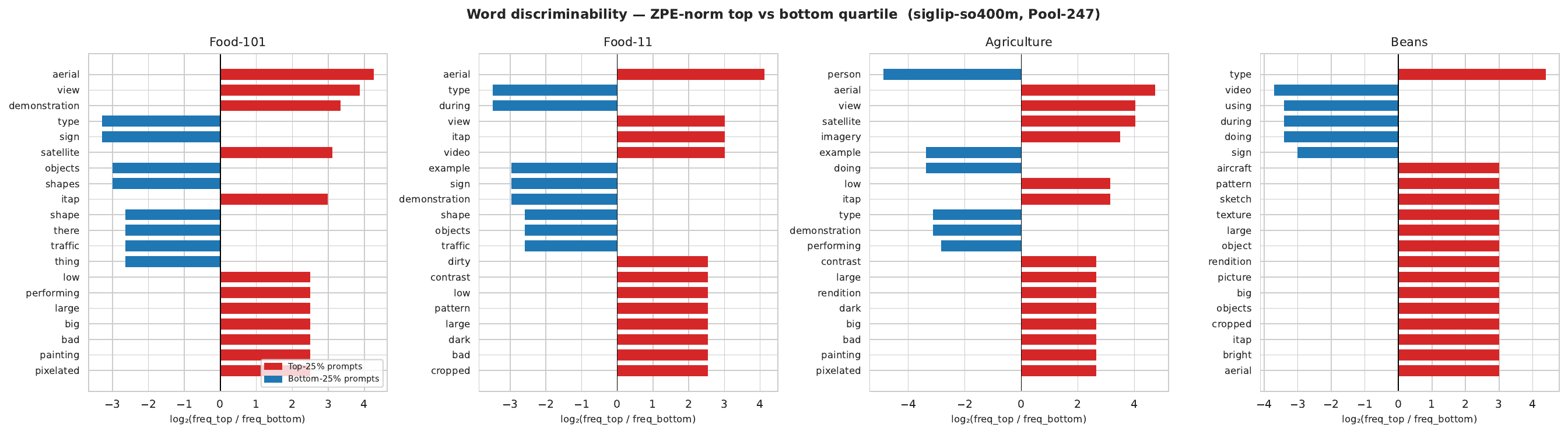}
   \caption{Word discriminability of top vs.\ bottom 
25\% ZPE-ranked prompts per dataset 
(SigLIP-SO400M, Pool-247).}
    \label{fig:word_discri}
\end{figure*}

\paragraph{Word-level discriminability.}
To better understand which lexical properties are associated with prompt quality, we compute a discriminability score for each word as the log-ratio of its frequency in the top-25\% versus bottom-25\% of prompts ranked by ZPE-norm score (see Fig.~\ref{fig:word_discri}). 
Regarding food-101 and Agriculture datasets, the words with the highest positive discriminability are mainly related to viewpoint and image framing, such as \textit{aerial}, \textit{view}, \textit{satellite} and \textit{overhead}. This suggests that prompts describing a concrete visual perspective tend to be more useful in these datasets. In contrast, words with negative discriminability are often less visually grounded. In Food-101, terms such as \textit{type}, \textit{sign}, \textit{shapes}, and \textit{objects} are associated with poorer prompts, while in Agriculture, low-scoring prompts often include words such as \textit{demonstration}, \textit{doing}, and \textit{performing}.

\begin{figure*}[t]
    \centering
    \includegraphics[width=1\linewidth]{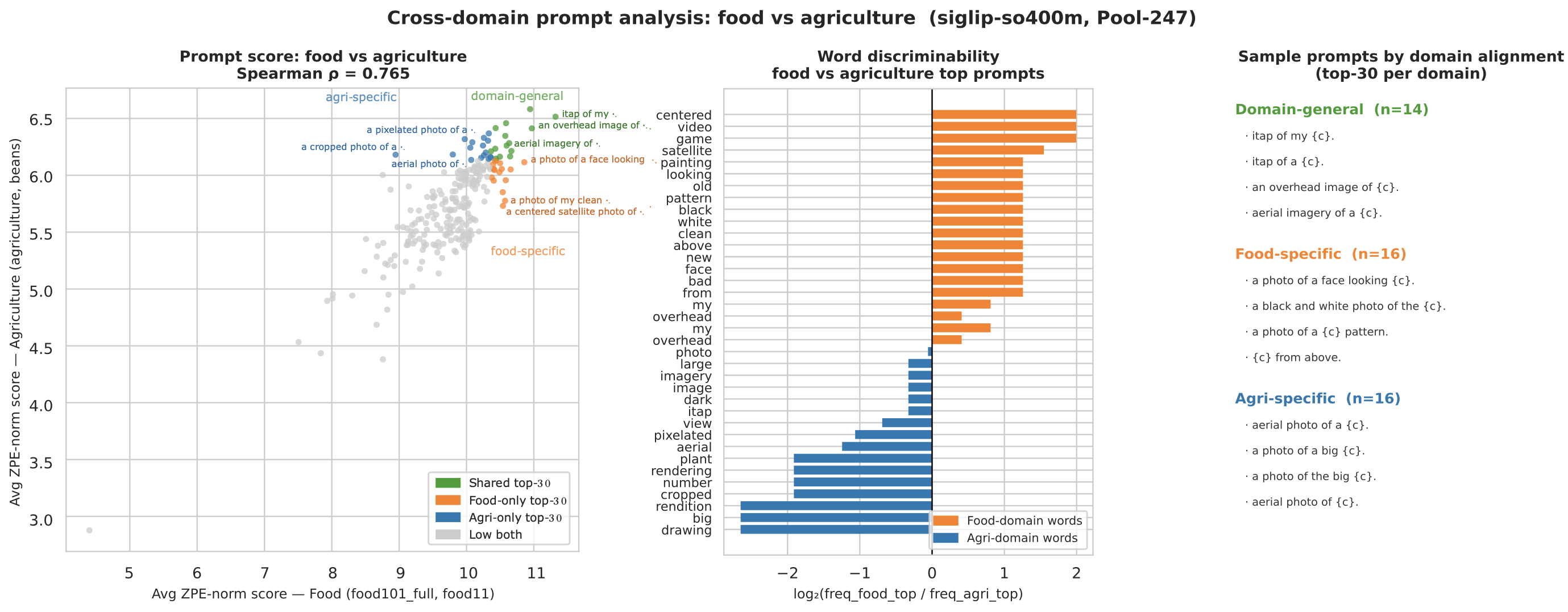}

    \caption{Cross-domain ZPE-norm prompt scores: 
Food vs.\ Agriculture (SigLIP-SO400M, Pool-247). 
$\rho = 0.765$ indicates strong cross-domain 
prompt quality transferability.}
    \label{fig:lexical}
\end{figure*}

\begin{figure*}[t]
    \centering
    \includegraphics[width=1\linewidth]{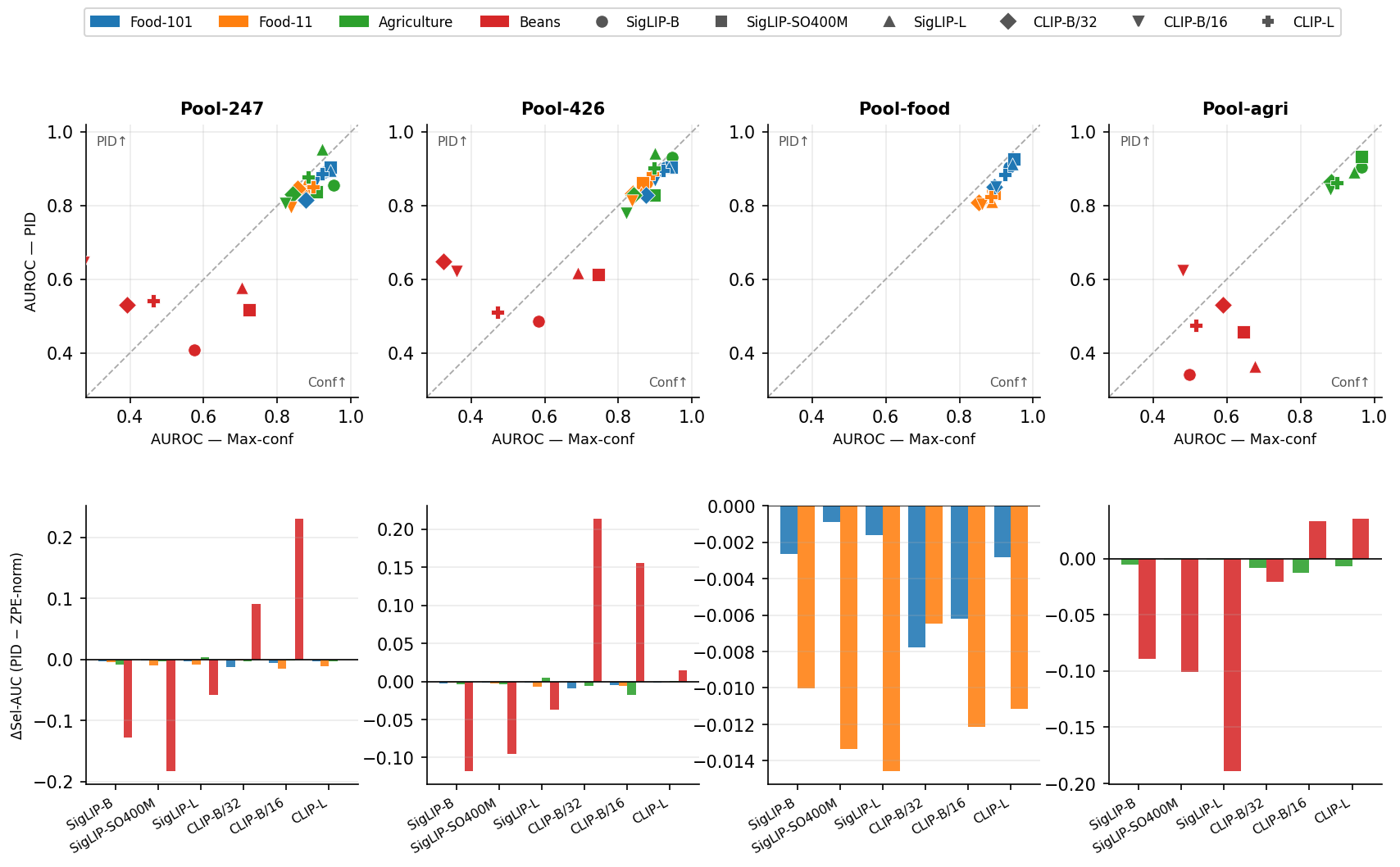}
    \caption{PID-based uncertainty evaluation across all prompt pools.}
    \label{fig:uncertaintly_all_pools}
\end{figure*}

\paragraph{Cross-domain effects: food vs.\ agriculture.}
Since Pool-247 is a generic pool, we also analyze whether ZPE-norm captures domain preferences when comparing Food-101 and Agriculture. Fig.~\ref{fig:lexical} plots the ZPE score of each prompt in the food domain against its score in the agriculture domain. The two score vectors are strongly correlated ($\rho=0.765$), indicating that many prompts  work well for both domains, which make sense, since they are relatively close topics.

In addition, prompts that are exclusive to the top-30 of each domain reveal a consistent lexical pattern. Food-specific prompts more frequently use definite constructions (e.g., ``aerial photo of \textbf{the} \{\}'' or ``satellite photo of \textbf{the} \{\}''), whereas agriculture-specific prompts tend to use indefinite forms (e.g., ``aerial photo of \textbf{a} \{\}'', ``aerial view of \textbf{a} \{\}''). While we do not explicitly model linguistic phenomena, this pattern suggests that prompt effectiveness may be sensitive to subtle grammatical choices (e.g., the/a). One possible explanation is that definite and indefinite forms induce different visual expectations, which may interact with how categories are represented across these two domains. Again, note that this effect is captured without access to domain labels.

Overall, this analysis indicates that ZPE does not only separate clearly poor prompts from useful ones, but also reflects more subtle lexical and domain-specific effects. 

\subsection{Uncertainty estimation and selective classification}

Fig. \ref{fig:uncertaintly_all_pools} shows the uncertainty as a proxy to detect prediction errors measured with AUROC. We compare both maximum confidence in the ensemble (max-conf) and PID.
\textcolor{blue}{Rather than a universal replacement for max-conf, we present PID as a complementary signal whose value is regime-dependent.} 
On ID datasets, (Food-101 and Food-11), max-conf is a strong baseline for
correctness prediction (AUROC 0.88--0.94 depending on model).
In this regime (where the model operates near its training
distribution) the model's own confidence already reflects when it will
fail. PID adds no substantial benefit here: the model already knows what it does not know. 

The picture changes sharply when the domain is far from pretraining, i.e., OOD datasets. On Agriculture, PID values overcome max-conf in several cases to detect errors.
On Beans, max-conf collapses as a predictor of correctness
(AUROC 0.27--0.72 depending on model).
The model's confidence no longer reflects visual evidence---it reflects
lexical biases in the embedding space.
PID recovers a useful signal: for CLIP-B/16 on Beans (Pool-247),
PID achieves AUROC\,=\,0.646 vs.\ max-conf AUROC\,=\,0.275, a gap of
$+0.37$ (the largest in the entire experiment).
The quality-gating mechanism is key: since CLIP has no representation of
plant diseases, no single prompt is reliably good.
But the \emph{collective} disagreement among all prompts, captured by a
large $\tau_a$, faithfully reflects the model's errors.

\subsection{$\tau_a$  as a domain diagnostic}
\begin{figure*}[h]
    \centering
    \includegraphics[width=1\linewidth]{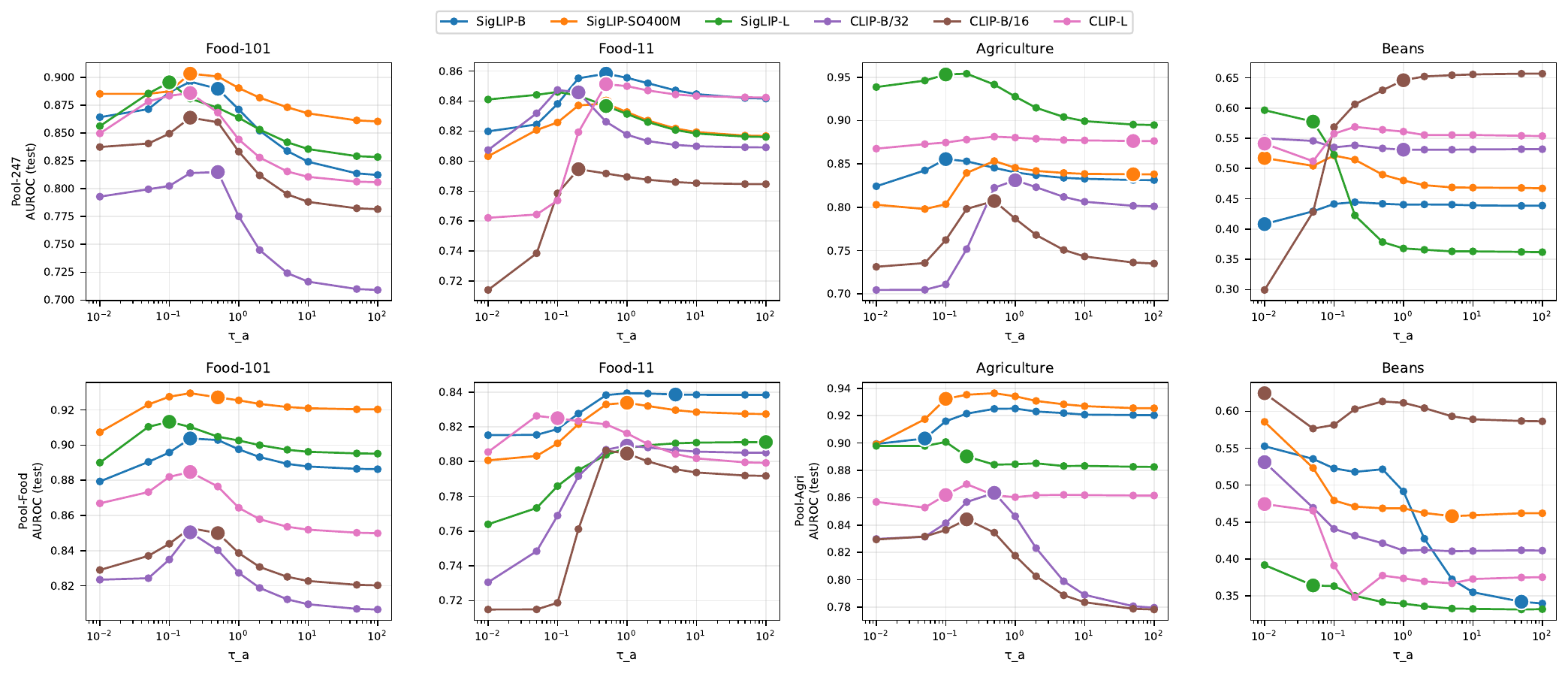}
\caption{PID AUROC as a function of $\tau_a$ 
across all datasets and prompt pools. Peaked 
curves (Food) indicate quality-gating is 
beneficial; monotonically increasing curves 
(Beans with CLIP) indicate full domain shift.}
    \label{fig:tau}
\end{figure*}

Fig.~\ref{fig:tau} shows PID AUROC as a function of $\tau_a$ for each
dataset.
The shape of this curve is itself informative:
\begin{itemize}
  \item \textbf{Peaked curve} (Food-101, $\tau_a^* \approx 0.2$--$0.5$):
        quality-gating helps.
        Best prompts define the disagreement signal; including all
        prompts equally introduces noise.
  \item \textbf{Monotonically increasing curve} (Beans with CLIP,
        $\tau_a^* = 1.0$, plateau at 100):
        no quality-gating needed.
        All prompts are equally weak, so collective disagreement is the
        only reliable signal; concentrating on the ``best'' prompts
        (which are no better than the rest) would discard information.
\end{itemize}
The $\tau_a$ curve serves as a \emph{domain-shift diagnostic}: a
peaked curve indicates the VLM has some grounding in the domain; a
monotonically increasing curve indicates domain shift.

\section{Conclusion}

We have studied zero-shot prompt ensembling in the agrifood domain and
shown that ZPE works as an unsupervised domain-alignment detector: without any labels, it assigns higher scores to prompts whose vocabulary
matches the visual domain.
This explains why small domain-specific pools consistently outperform large generic ones, and directly motivates PID, a
quality-gated uncertainty method that repurposes prompt disagreement to
predict model failures.

PID is used to measure the epistemic uncertainty in the disagreement, incorporating a hyperparameter to only compute the disagreement in high-quality prompts. We use this uncertainty for selective classification, finding that PID estimates errors reliably and competitively with max-conf under mild shift, and provides tighter error bounds under severe domain shift.
The $\tau_a$ ablation curve serves as a practical diagnostic: its shape
reveals if the model is domain-grounded or is operating
under full domain shift.

These contributions provide a more complete picture of when and
why prompt ensembling works in agrifood domains, and offer practical guidance for prompt pool
design in specialised visual domains, and how quality-gating impacts performance, calibration, and selective prediction.

\paragraph{Future work.} There are still several open questions that we may address in future work. For example, why do semantically weak prompts score high?
An interesting observation is that some prompts with seemingly weak or
unrelated semantics still appear among high-scoring prompts.
This suggests that prompt effectiveness is not solely driven by explicit
semantic alignment.
One possibility is that they capture indirect visual cues or regularization
effects in the model's representation space.
Understanding these mechanisms remains an open question for future work. In another vein, the role of PID has to be studied more in-depth in a wider setting to fully unveil, when epistemic uncertainty derived from disagreement can improve max-conf as a proxy for reliability in the outcomes.
\textcolor{blue}{Finally, our analysis is restricted to English, given the predominantly English-trained encoders used here. Extending it to morphologically richer languages such as Spanish, where phenomena like grammatical gender and noun--adjective agreement may affect prompt consistency, is a promising direction, requiring multilingual VLMs and designing Spanish prompt pools.}

\paragraph{Limitations.} Optimal $\tau^*$, $\tau_a^*$ and  $c$ values are fitted on validation
data, which deviates from a fully zero-shot setting.
Future work could explore data-free approaches to temperature selection.
Additionally, the Beans dataset shows near-random performance with CLIP
models, likely due to its small size, class imbalance, and the large
domain gap from pretraining; this makes it a useful stress-test but limits
conclusions about PID on larger out-of-domain benchmarks.
We also assume that food-related data is more extensively represented in pre-trained VLMs due to the wide availability of food-related content on the web and in training datasets, in contrast to agriculture, which remains a less explored domain. Further large-scale studies are required to validate this assumption.

\section*{Acknowledgments}
Work supported by the FederaTrans project: Grant PID2024-158373OB-I00 funded by MICIU/AEI/10.13039/501100011033 and
by ERDF/EU. It is also funded by the European Union (BAG-INTEL project, grant agreement no. 101121309 and CUSTOMAI project, grant
agreement no. 101226029).

\bibliographystyle{fullname_esp}
\bibliography{EjemploARTsepln}

\end{document}